\documentclass[11pt]{article}

\usepackage{coling2018}
\usepackage{times}
\usepackage{latexsym}
\usepackage{graphicx}
\usepackage{amsmath}
\usepackage{amsfonts,amssymb}
\usepackage{subfigure}
\usepackage{url}

\title{An Annotation Scheme of A Large-scale Multi-party Dialogues Dataset for Discourse Parsing and Machine Comprehension}
\author{Jiaqi Li, Ming Liu, Bing Qin, Zihao Zheng, Ting Liu\\
Harbin Institute of Technology\\
\{jqli, mliu, qinb, zhzheng, tliu\}@ir.hit.edu.cn}
\begin{document}
\maketitle

\begin{abstract}

In this paper, we propose the scheme for annotating large-scale multi-party chat dialogues for discourse parsing and machine comprehension. 
The main goal of this project is to help understand multi-party dialogues. %In detail, this manual contains the motivation for the corpus and the specific content in the corpus. 
Our dataset is based on the Ubuntu Chat Corpus. For each multi-party dialogue, we annotate the discourse structure and question-answer pairs for dialogues. As we know, this is the first large scale corpus for multi-party dialogues discourse parsing, and we firstly propose the task for multi-party dialogues machine reading comprehension.

\end{abstract}

\section{Introduction}

% 介绍现状和研究意义
There are more and more NLP scholars focusing on the research of multi-party dialogues, such as multi-party dialogues discourse parsing and multi-party meeting summarization \cite{shi2019deep,GSNijcai2019,li2019keep,zhao2019abstractive,N16IntegerLP,D15DP4MultiPparty}. However, the scale of the STAC dataset has limited the research of discourse parsing for multi-party dialogues. On the other hand, as we know, there is no literature working on machine reading comprehension for multi-party dialogues. Considering the connection between the relevance between machine reading comprehension and discourse parsing, we annotate the dataset for two tasks for multi-party dialogues understanding.

Our dataset derives from the large scale multi-party dialogues dataset the Ubuntu Chat Corpus \cite{lowe2015ubuntu}. For each dialogue in the corpus, we annotate the discourse structure of the dialogue and propose three questions and find the answer span in the input dialogues. %Our dataset contains 38K dialogues and 1.75M utterances. 
To improve the difficulty of the task, we annotate $ \frac{1}{6}$ to $ \frac{1}{3}$ unanswerable questions and their plausible answers from dialogues.

This is a real example from the Ubuntu dataset.

%Figure 1 shows a real dialogue in our dataset. There are four speakers and nine utterances in the sample dialogue. The left part shows the speakers and their utterances and the right part shows the discourse dependency relation arcs. The discourse structure can be seen as a discourse dependency graph. We adopt the same sense hierarchy with the STAC dataset which contains sixteen discourse relations. 

\textbf{Example 1}\\
\fbox{%
	\parbox{0.95\textwidth}{%
		1. mjg59: Someone should suggest to Mark that the best way to get people to love you is to hire people to work on reverse-engineering closed drivers. \\
		2. jdub: heh  \\ 
		3. daniels $\rightarrow$ mjg59: heh \\
		4. daniels: HELLO \\
		5. daniels $\rightarrow$ mjg59: your job is to entertain me so I don't fall asleep at 2pm and totally destroy my migration to AEST \\
		6. bdale $\rightarrow$ daniels: see you next week? \\
		7. daniels $\rightarrow$ bdale: oh, auug, right. rock. \\
		8. daniels $\rightarrow$ bdale: just drop me an email, or call +61 403 505 896 \\
		9. bdale $\rightarrow$ daniels: I arrive Tuesday morning your time, depart Fri morning, will be staying at the Duxton \\
	}
}\\

There are mainly two contributions to our corpus:

\begin{itemize}
	\item A first large scale multi-part dialogues dataset for discourse parsing. It is a challenging task to parse the discourse structure of multi-party dialogues. Enough training data will be essential to develop more powerful models.
	\item We firstly propose the task of machine reading comprehension for multi-party dialogues. Different from existing machine comprehension tasks, multi-party dialogues could be more difficult which needs a graph-based model for representing the dialogues and better understanding the discourse structure in the dialogue.
\end{itemize}

In this paper, I will give a detailed description of our large scale dataset. In section 2, I will introduce Ubuntu corpus. In Section 3 and Section 4, I will introduce the annotation for discourse parsing and machine reading comprehension respectively. In Section 5, I will briefly list some related literature.

\section{Ubuntu Corpus}

Our dataset derives from the large scale multi-party dialogues dataset the Ubuntu Chat Corpus \cite{lowe2015ubuntu}. The Ubuntu dataset is a large scale multi-party dialogues corpus. 

There are several reasons to choose the Ubuntu dataset as our raw data for annotation.

\begin{itemize}
    \item First, Ubuntu dataset is a large multi-party dataset. Recently, \cite{GSNijcai2019} used Ubuntu as their dataset for learning dialogues graph representation. After some preprocessing, there are 38K sessions and 1.75M utterances. In each session, there are 3-10 utterances and 2-7 interlocutors.
    \item Second, it is easy to annotate the Ubuntu dataset. The Ubuntu dataset already contains Response-to relations that are discourse relations between different speakers' utterances. For annotating discourse dependencies in dialogues, we only need to annotate relations between the same speaker's utterances and the specific sense of discourse relation. %Because the length of dialogues in Ubuntu dataset is not too long, we can easily to summarize dialogues and propose some questions for the dialogue.  
    \item Third, there are many papers doing experiments on the Ubuntu dataset, and the dataset has been widely recognized.
\end{itemize}

The discourse dependency structure of each multi-party dialogue can be regarded as a graph. To learn better graph representation of multi-party dialogues, we adopt the dialogues with 8-15 utterances and 3-7 speakers. To simplify the task, we filter the dialogues with long sentences (more than 20 words). Finally, we obtain 52,053 dialogues and 460,358 utterances.

\section{Annotation for discourse parsing in multi-party dialogues}

This section will explain how to annotate discourse structure in multi-party dialogues. 

The task of discourse parsing for multi-party dialogues aims to detect discourse relations among utterances. The discourse structure of a multi-party dialogue is a directed acyclic graph (DAG). In the process of annotation of discourse parsing for multi-party dialogues, there are two parts: edges annotation between utterances and specific sense type of discourse relations.

The discourse structure of Example 1 is shown in Figure 1. There are four speakers and nine utterances in the sample dialogue. The left part shows the speakers and their utterances and the right part shows the discourse dependency relation arcs. The discourse structure can be seen as a discourse dependency graph. We adopt the same sense hierarchy with the STAC dataset which contains sixteen discourse relations.

\begin{figure}
	\centering
	\includegraphics[width=0.6\textwidth]{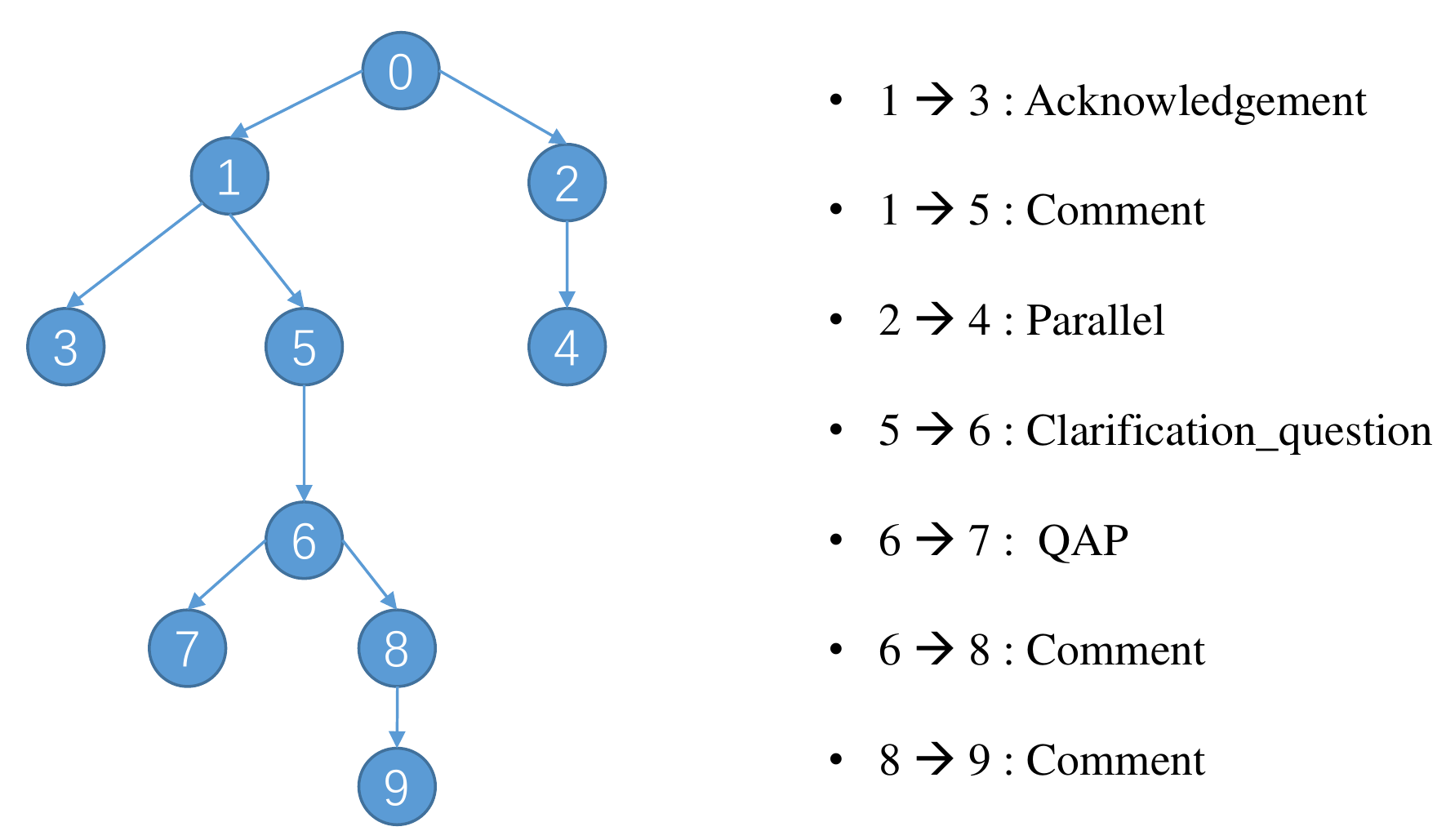}
	\caption{ The discourse dependency structure and relations for \textbf{Example 1}. }
	\label{fig:framework}
\end{figure}

\subsection{Edges between utterances}

The edge between two utterances represents that there is the discourse dependency relations between these two utterances. The direction of the edge represents the direction of discourse dependency. In this subsection, what we need to do is to confirm whether two utterances have discourse relation. Like PDTB \cite{prasad2008penn}, we call two utterances as Arg1 and Arg2 respectively. The front utterance is Arg1 and the back utterance is Arg2.

For example, there is a multi-party dialogue with 9 utterances in Example 1, utterances 1-9 respectively. The utterance 3 depends on utterance 1, we can draw an edge from utterance 1 to utterance 3. Otherwise, if utterance 1 depends on utterance 2, we can draw an edge from utterance 2 to utterance 1. In most cases, the direction of discourse relations in multi-party dialogues is from the front to the back.

The biggest difference between discourse parsing for well-written document and dialogues is that discourse relations can exist on two nonadjacent utterances in dialogues. When we annotate dialogues, we should read dialogues from begin to the end. For each utterance, we should find its one parent node at least from all its previous utterances. We assume that the discourse structure is a connected graph and no utterance is isolated. 

\subsection{Sense of discourse relations}

When we find the discourse relation between two utterances, we need continue to confirm the specific relation sense. We adopt the same senses hierarchy with the STAC dataset. 

There are sixteen discourse relations in the STAC. All relations are listed as follows: Comment, Clarification\_question, Elaboration, Acknowledgement, Continuation, Explanation, Conditional, Question-answer\_pair, Alternation, Q-Elab, Result, Background, Narration, Correction, Parallel, Contrast.

\section{Annotation for machine reading comprehension in multi-party dialogues}

The task of reading comprehension for multi-party dialogues aims to be beneficial for understanding multi-party dialogues. Different from existing machine reading comprehension tasks, the input of this task is a multi-party dialogue, and we should to answer some questions given the dialogue. 

We propose three questions for eache dialogue and annotate the span of answers in the input dialogue. As we know, our dataset is the first corpus for multi-party dialogues reading comprehension. 

We construct following questions and answers for the dialogue in \textbf{Example 1}:

\begin{itemize}
\item Q1: When does Bdale leave?
\item A1: Fri morning
\item Q2: How to get people love Mark in Mjg59's opinion.
\item A2: Hire people to work on reverse-engineering closed drivers.
\end{itemize}

On the other hand, to improve the difficulty of the task, we propose $ \frac{1}{6}$ to $ \frac{1}{3}$ unanswerable questions in our dataset. We annotate unanswerable questions and their plausible answers (PA). Each plausible answer comes from the input dialogue, but is not the answer for the plausible question. 

\begin{itemize}
	\item Q1: Whis is the email of daniels?
	\item PA: +61 403 505 896
\end{itemize}

\section{Related work}

% explicit discourse parsing.

%In explicit discourse relation recognition, due to the fact that connectives can indicate discourse relation, recent methods got good performance. Pitler et al used an unsupervised method and got a good result only using the connective \cite{pitler2009using}. Besides, there are some supervised methods to recognize the explicit discourse relation. For instance, Pitler et al used an approach based on some syntactic features related to the connective, and got an improvement in explicit discourse relation recognition. To reduce errors propagation, a joint learning approach via structured perceptron for explicit discourse parsing was proposed, they got comparable results on relation classification and got an improvement on argument labeling \cite{li2014joint}.

%\subsection{Problem Definition}
%
%We formulate the task of discourse parsing on multi-party chat dialogues as follows:
%
%\textbf{Input}: $D = \{u_0, u_1, u_2,..., u_n\}$,  where $D$ is the sequence of EDUs in a multi-party chat dialogue with $n$ messages, and $u_i$ is the ith message in the dialogue.
%
%\textbf{Output}: $G(V,E,R)$, where $V$ represents vertex set consists of EDUs and $|V|=n$, and $E$ represents edge set between EDUs, and $R$ represents discourse relations.

In this section, I will introduce several existing multi-party dialogues datasets, and explain why we need to annotated a new dataset.

\subsection{Discourse parsing for multi-party dialogues}

There is an only corpus of discourse parsing on multi-party chat dialogues: STAC \cite{asher2016discourse}. The corpus derives from online game \textit{The Settlers of Catan}. The game \textit{Settlers} is a multi-party, win-lose game. As mentioned above, an example in STAC is shown in Figure 1. More details for STAC corpus are described in \cite{asher2016discourse}.

The overview of the STAC is shown in Table 1. From Table 1 we can know that there are about more 10K EDUs and relations and most of EDUs are weakly connected. Each EDU can be regarded as a message or sentence in the dialogues.

% 实验数据规模
\begin{table}[h]
	\begin{center}
		\begin{tabular}{l|r|l|c}
			\hline \bf   \bf  & \bf Total & \bf Training  & \bf Testing \\ 
			\hline
			Dialogues   & 1091   & 968   & 123   \\
			EDUs   & 10677   & 9545   & 1132    \\ 
			Relations   & 11348   & 10158   & 1190   \\
			\hline
		\end{tabular}
	\end{center}
	\caption{\label{font-table} Overview of STAC corpus.}
\end{table}

There are sixteen types of discourse dependency relations in STAC as shown in Section 3.2.

% 实验数据规模
%\begin{table}[h]
%	\begin{center}
%		\begin{tabular}{l|r|l|c}
%			\hline \bf   \bf  & \bf Total & \bf Training  & \bf Testing \\ 
%			\hline
%			Comment   & 1851   & 1684   & 167   \\
%			Clarification\_question   & 260   & 240   & 20    \\ 
%			Elaboration   & 869   & 771   & 98   \\
%			Acknowledgement   & 1010   & 893   & 117   \\
%			Continuation   & 987   & 10158   & 1190   \\
%			Explanation   & 437   & 407   & 30   \\
%			Conditional   & 124   & 105   & 19   \\
%			Question-answer\_pair   & 2541   & 2236   & 305   \\
%			Alternation   & 146   & 128   & 18   \\
%			Q-Elab   & 599   & 525   & 74   \\
%			Result   & 578   & 551   & 27   \\
%			Background   & 61   & 58   & 3   \\
%			Narration   & 130   & 116   & 14   \\
%			Correction   & 212   & 189   & 23   \\
%			Parallel   & 215   & 196   & 19   \\
%			Contrast   & 493   & 449   & 44   \\
%			\hline
%			TOTAL   & 10513   & 9421   & 1092   \\
%			\hline
%		\end{tabular}
%	\end{center}
%	\caption{\label{font-table} Distribution of discourse relations in STAC corpus \cite{rehbein2016STAC}.}
%\end{table}

\subsection{Machine reading comprehension}

Machine reading comprehension is a popular task which aims to help the machine better understand natural language. There are several types of datasets for machine comprehension, including extractive datasets \cite{joshi2017triviaqa,trischler2017newsqa}, answer sentence selection datasets \cite{wang2007jeopardy,yang2015wikiqa} and multiple choice datasets \cite{richardson2013mctest,lai2017race}. I will briefly introduce two datasets QuAC \cite{choi2018quac}and CoQA \cite{reddy2019coqa}.

%MCTest、Race、SQuad、SQuad 2.0

QuAC : Question Answering in Context is a two-party dialogues dataset for machine reading comprehension \cite{choi2018quac}. The dataset for Question Answering in Context that contains 14K information-seeking QA dialogs (100K questions in total). The dialogs involve two crowd workers: (1) a student who poses a sequence of freeform questions to learn as much as possible about a hidden Wikipedia text, and (2) a teacher who answers the questions by providing short excerpts from the text.

CoQA is a large dataset for building conversation question answering systems \cite{reddy2019coqa}.

\section{Conclusion}

We propose the scheme for annotating large scale multi-party chat dialogues for discourse parsing and machine comprehension. 
The main goal of this project is to be beneficial for understanding multi-party dialogues. %In details, this manual contains the motivation for the corpus and the specific content in the corpus. 
Our corpus are based on the Ubuntu Chat Corpus. For each multi-party dialogue, we annotate discourse structure and question-answer pairs for the dialogue. As we know, this would be the first large-scale corpus for multi-party dialogues discourse parsing, and we firstly propose the task for multi-party dialogues machine reading comprehension.

%In this document, I describe the research work that I have done and some future research plans. The first section includes the annotation of HIT-CDTB 2.0, implicit discourse relation recognition with structured attention in PDTB, and unsupervised Chinese style transfer. The second section lists several preliminary research plans about discourse structure and its application.

%There are several research work that ...

\bibliography{statement}
\bibliographystyle{acl}

\end{document}